\documentclass[letterpaper, 10 pt, conference]{ieeeconf}  

\IEEEoverridecommandlockouts                              
\usepackage{graphicx}      
\usepackage{algorithm} 
\usepackage{algpseudocode} 
\usepackage{varwidth} 
\usepackage{cite}
\usepackage{diagbox}
\usepackage{subcaption}
\usepackage{amsfonts}
\usepackage{amsmath}
\usepackage{url}
\usepackage{gensymb}
\usepackage{tabularx,caption}
\usepackage[dvipsnames]{xcolor}

\graphicspath{{figs/}}
\title{\LARGE \bf
Learning to Synthesize Volumetric Meshes from \\ Vision-based Tactile Imprints
}

\author{Xinghao Zhu$^{1,2}$, Siddarth Jain$^2$, Masayoshi Tomizuka$^1$, and Jeroen van Baar$^2$
\thanks{$^1$ Mechanical Systems Control Lab, UC Berkeley, Berkeley, CA, USA. 
{\tt\small \{zhuxh,tomizuka\}@berkeley.edu}}
\thanks{$^2$ Mitsubishi Electric Research Laboratories (MERL), Cambridge, MA, USA
{\tt\small {\{sjain,jeroen\}@merl.com}}
}}

\begin{document}
\maketitle
\thispagestyle{empty}
\pagestyle{empty}

\begin{abstract}
Vision-based tactile sensors typically utilize a deformable elastomer and a camera mounted above to provide high-resolution image observations of contacts. Obtaining accurate volumetric meshes for the deformed elastomer can provide direct contact information and benefit robotic grasping and manipulation. This paper focuses on learning to synthesize the volumetric mesh of the elastomer based on the image imprints acquired from vision-based tactile sensors. Synthetic image-mesh pairs and real-world images are gathered from 3D finite element methods (FEM) and physical sensors, respectively. A graph neural network (GNN) is introduced to learn the image-to-mesh mappings with supervised learning. A self-supervised adaptation method and image augmentation techniques are proposed to transfer networks from simulation to reality, from primitive contacts to unseen contacts, and from one sensor to another.
Using these learned and adapted networks, our proposed method can accurately reconstruct the deformation of the real-world tactile sensor elastomer in various domains, as indicated by the  quantitative and qualitative results.
\end{abstract}


\section{Introduction}
\label{sec: introduction}

\begin{figure}[tb]
\begin{center}
	\includegraphics[width=3.4in]{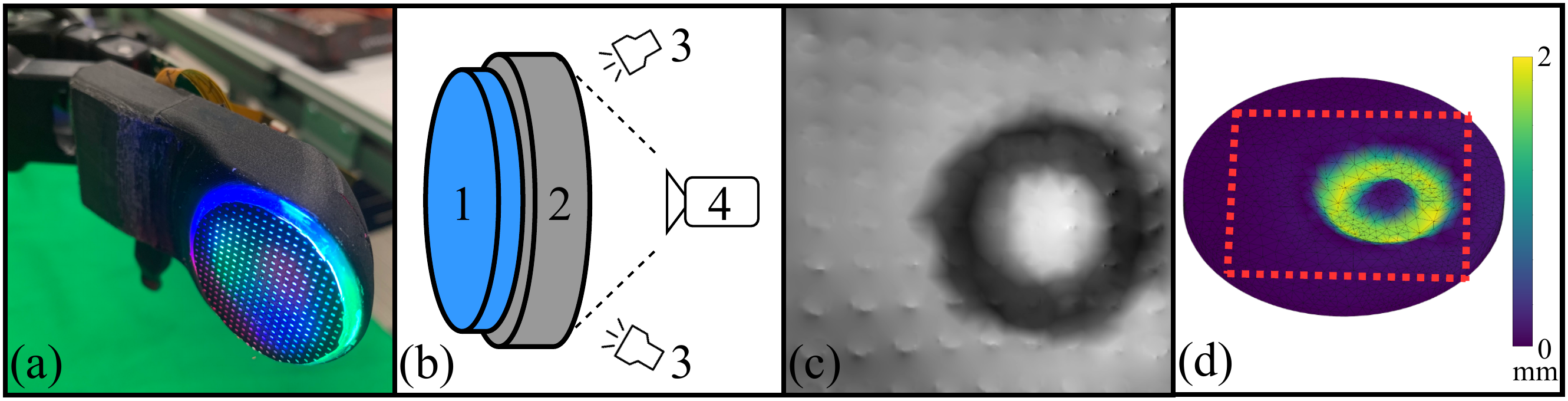}
	\caption{\textbf{(a)} the GelSlim visual-tactile sensor,  \textbf{(b)} the construction of the sensor, with the elastomer (1), the transparent lens (2), the lights (3), and the camera (4).
	\textbf{(c)} a depth image observation obtained from the sensor, and \textbf{(d)} the corresponding reconstructed volumetric mesh with our method. 
	The red rectangle denotes the camera's view range, and the color represents the displacement level.}
	\label{fig: teaser}
\end{center}
\end{figure}

Tactile is an essential sensing modality for humans when grasping and manipulating objects.
Tactile sensors can provide direct information about contacts during robotic grasping and manipulation. Vision-based tactile sensors are variants among different designs for robotic tactile sensors~\cite{gelsight, gs1, gs2, gs3, stretch, SOFTcell, lambeta2020digit, padmanabha2020omnitact}. These sensors use a camera to capture high spatial resolution images of the contact deformation of a piece of elastomeric gel with an opaque coating as the sensing surface, as shown in Fig.~\ref{fig: teaser} (a) and (b). 

Obtaining a mesh representation of the contact elastomer can advance the development of applications with vision-based tactile sensors, since meshes can provide accurate contact information. For instance, meshes of the elastomer have enabled in-hand object localization~\cite{8794298, narang2020interpreting, bauza2020tactile}, vision-free manipulation~\cite{tactilerl,9196976,she2020rss}, and contact profile reconstruction~\cite{gs2,gs3,8641397,extrinsic_contact_sense, swingbot}. Also, meshes can be used for precise dynamics simulation~\cite{mesh_based_sim, gnn_sim, gnn_complex_physics} and future state estimation~\cite{belbute_peres_cfdgcn_2020, li2020visual}.


Previous simulation studies~\cite{gs2,gs3} for vision-based sensors focus on reconstructing the \textit{surface mesh} by tracking markers on the sensor.  This can provide the surface displacement fields of the elastomer. However, to better simulate the dynamics, a \textit{volumetric mesh} is preferred~\cite{gnn_sim}. Compared to the \textit{surface mesh}, the \textit{volumetric mesh} contains internal vertices and edges, thus can better encode the dynamics and estimate the contact profile with the Finite Element Method (FEM)~\cite{fembook,fembook2,mesh_based_sim}. Nevertheless, internal elements also challenge the reconstruction of the \textit{volumetric mesh} due to additional dimensions. This paper addresses that challenge and proposes a method to directly predict the \textit{volumetric mesh} from images using vision-based tactile sensors, such as the GelSlim~\cite{gs3}, in a sim-to-real setting. Moreover, our approach does not rely on fiducial sensor markers to synthesize a \textit{volumetric mesh}.


We first employ 3D FEM simulations of the GelSlim sensor's elastomer to collect image-mesh data pairs. The FEM simulations compute volumetric deformation fields for the elastomer with arbitrary contacts. The depth image observation is then rendered with synthetic cameras. The contact experiments are also executed in the real world with physical GelSlim. However, real-world contacts only provide images, since ground-truth meshes are unprocurable. We then learn mappings from real-world images to mesh deformations (as shown in Fig.~\ref{fig: teaser} (c) and (d)) by leveraging supervised pre-training and self-supervised adaptations. Specifically, we learn an image-to-mesh projection in latent space with synthetic data pairs. 

Sim-to-real approaches have to overcome the distribution differences between the two domains, that is the sim-to-real gap. We propose data augmentation of the synthetic images together with a self-supervised adaptation method on real-world images to address this gap. The adaptation uses a differentiable renderer to project the network output into images and minimize the difference between projected and input images. We demonstrate that this adaptation can transfer networks for sim-to-real, seen contact objects to novel contact objects, and between different GelSlim sensor instances. In this paper our goal is to introduce the synthesis of volumetric meshes from tactile imprints, and will address applications with our approach in the future work.


Our work makes the following contributions:
\begin{itemize}
  \item We provide a FEM model for GelSlim tactile sensors with a GPU-based simulator and propose a method to calibrate the FEM model with physical GelSlim sensors.
  \item We collect contact datasets from synthetic and real-world contact experiments for GelSlim sensors.
  \item We present an image-to-mesh projection network to reconstruct the volumetric mesh of the elastomer without the need for fiducial sensor markers.
  \item We further propose a self-supervised adaptation method and image augmentation techniques to mitigate the domain shift of sensor readings.
\end{itemize}

The problem formulation and details of our method are described in Section~\ref{sec: algorithms}, followed by experimental results in Section~\ref{sec: exp_results}, and a discussion in Section~\ref{sec: discussion}. We discuss the related work in the next section.



\section{Related Work}
\label{sec: related_works}

A variety of tactile sensing capabilities for robotic applications have been introduced recently~\cite{li2020review, yamaguchi2019recent}. In this paper, we limit our discussion to vision-based sensors, such as GelSight~\cite{gelsight}, and GelSlim~\cite{gs1}.


It is nontrivial to convert the acquired tactile sensory information to quantities relevant for performing robotic tasks, such as the grasping force. While data-driven approaches to tactile sensing are becoming more popular, collecting large real-world datasets is not feasible due to expense and potential for damage. A promising solution is to investigate sim-to-real methods for tactile sensing capabilities. 


Typically tactile sensors rely on soft materials, such as elastomers, wherein the contact results in deformation of the material. For vision-based tactile sensors, simulation thus involves modeling both visual and deformation behavior. Visual output of the GelSight sensor was simulated in~\cite{gomes2021generation} using a depth camera in the Gazebo simulator~\cite{koenig2004design}. The approach involves computing the heightmap of the elastomer from the depthmap, and approximating the internal illumination of the elastomer using a calibrated Phong’s reflection model. The illumination stage approach is further extended in~\cite{wang2020tacto} by leveraging OpenGL~\cite{opengl}. In our approach, we do not only attempt to bring the simulated tactile image closer to the real world, but we also use a data-driven approach with self-supervised adaption to map from tactile images to volumetric meshes. Furthermore, our approach has the potential to help with the need for texture augmentation, important in real-world robotic applications.


Analytical modeling with FEM simulation can model contact dynamics~\cite{fembook}. This has been used in the analysis of the behavior of soft materials for tactile sensing under various conditions. A variety of simulation methods have been considered for different types of tactile sensors using the FEM~\cite{gs2, sferrazza2019ground, narang2020interpreting, biotac_sim2real}. 
 
The elastomer of the GelSlim tactile sensor is modeled as a linear elastic material in~\cite{gs2}, and the FEM is used to compute the stiffness matrix to approximate its external forces and displacements. Using the surface displacements, this matrix is then used to compute an estimate of the force distribution. Unlike~\cite{gs2}, in our approach we don't require the use of fiducial tracking markers to determine the displacement of the elastomer. Furthermore, we estimate volumetric meshes directly, which is not explored in  prior work. Our approach is akin to the FEM model of the SynTouch BioTac sensor~\cite{biotac_sim2real}, in that both learn latent representations for the simulated sensor deformations and the real-world output through self-supervision. In contrast to~\cite{biotac_sim2real}, we focus on the image to volumetric mesh projection for vision-based tactile sensors. Since image observations from the sensor have higher variance and are noisy, our problem is more challenging.

 

\section{Methods}
\label{sec: algorithms}

This section first introduces the problem statement and preliminaries. Next, the image-to-mesh projection and self-supervised adaptation methods are discussed. Finally, the datasets are described, including synthetic labeled data, real-world unlabeled data, and the data augmentation techniques.

\subsection{Problem Statement and Preliminaries}
This paper focuses on the problem of reconstructing an elastomer's volumetric mesh with image observations for vision-based tactile sensors. The non-injective projection (or mapping) from surface images to volumetric vertex positions makes this problem nontrivial. Some preliminaries are described below:

\subsubsection{Image Observations}
\label{subsubsec: img_obs}
Visual tactile sensors typically contact objects with a silicone elastomer and use a camera to capture the deformation of the surface, as shown in Fig~\ref{fig: teaser}.
The captured RGB image can be used to construct a depth map of the contact surface using shape from shading~\cite{depth_recons, gs3}. It establishes a mapping from the RGB color to the surface normals with a marble of known dimension. During runtime, surface normals are retrieved and integrated into the depth map $I$. Compared to raw RGB images, depth maps contain 2.5D information and can better represent the geometry of the contact surface~\cite{cgpn}. Moreover, depth maps are much easier to simulate using synthetic cameras and thus have less sim-to-real gap. Therefore, in this paper we use (128$\times$128) depth maps $I$ as the image observations.


\subsubsection{Volumetric Meshes with FEM}
The FEM is a mathematical tool to solve complex partial differential equations (PDEs)~\cite{fembook}. 
In the FEM, geometrical shapes are represented by volumetric meshes $\mathcal{M}$, which consist of 3D elements, such as tetrahedrons and hexahedrons. With high-resolution meshes and small computation steps, FEM can estimate the forward dynamics of soft bodies~\cite{fembook2,mesh_based_sim}.

This paper uses graphs to represent volumetric meshes. Specifically, volumetric meshes are defined as a set of vertices and edges, $\mathcal{M} = (\mathcal{V}, \mathcal{A})$, with $n$ vertices in 3D Euclidean space, $\mathcal{V} \in \mathbb{R}^{n \times 3}$. The adjacency matrix $\mathcal{A} \in \{ 0, 1 \}^{n \times n}$ represents the edges. If vertices $i$ and $j$ are connected by an edge,  $\mathcal{A}_{ij}=1$, and $\mathcal{A}_{ij}=0$ otherwise.


\begin{figure}[tb]
\begin{center}
	\includegraphics[width=3.4in]{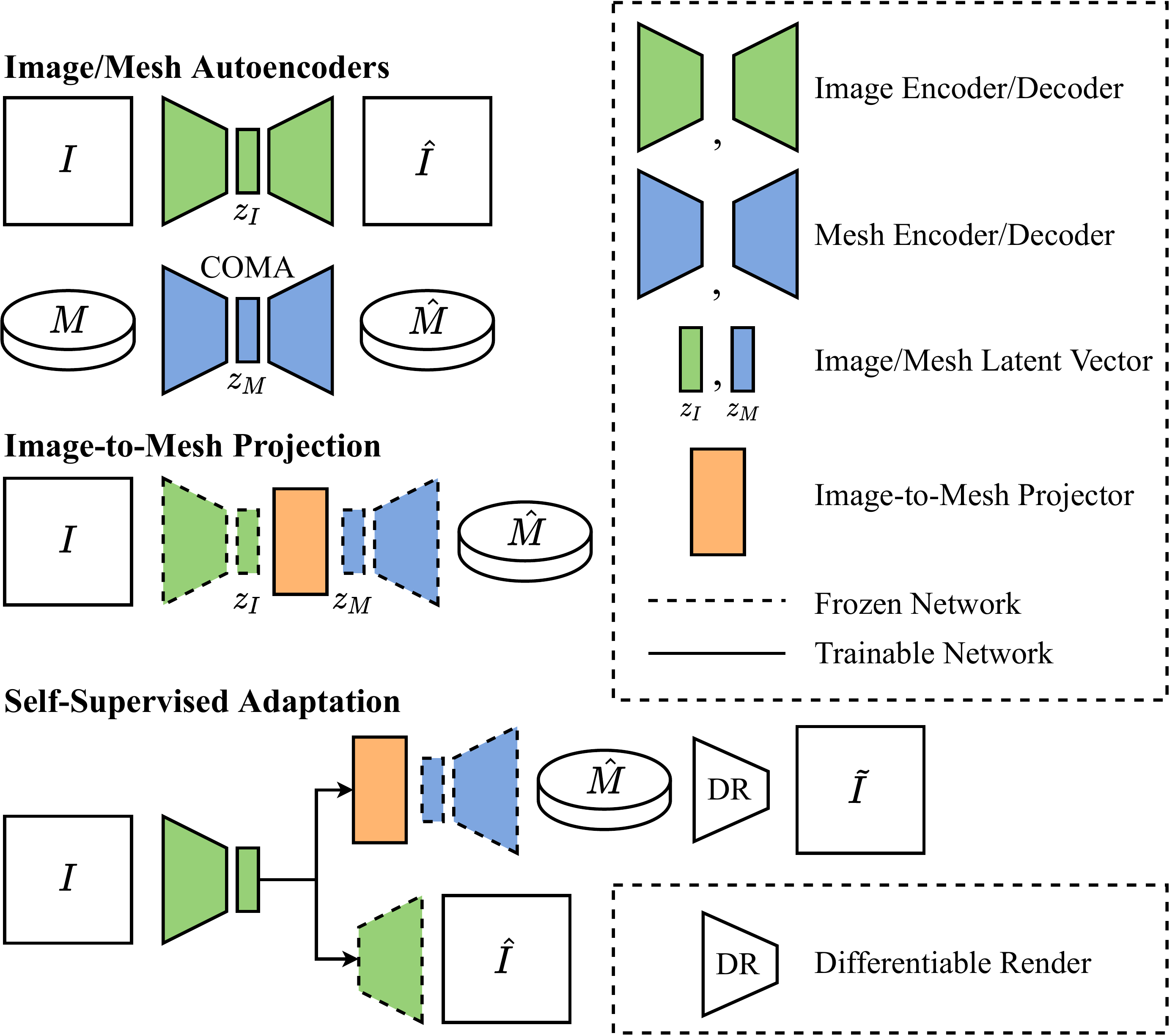}
	\caption{Training structure. The image-to-mesh projection network is optimized with pre-trained autoencoders. The self-supervised adaptation transfers the projection network to various domains with a differentiable render. }
	\label{fig:architecture}
\end{center}
\end{figure}

\subsection{Supervised Image-to-Mesh Projection}
\label{subsec: i2m}
Our goal is to map an input depth map $I$ to a volumetric mesh $\mathcal{M}$.
Although depth maps provide geometrical information for the contact surface, the projection from surface images to volumetric vertex positions is not injective and is hard to analyze.
Specifically, different displacements can generate the same surface observation.
Thus, in this paper we assume a fixed mesh tessellation (i.e., $\mathcal{A}$ fixed) to enforce the injective mapping and use a neural network to learn the underlying projection $\hat{\mathcal{M}} = f_\theta(I)$, with $\theta$ being the parameters of the network.

The image-to-mesh projection is learned with latent representations.
Compared to previous work~\cite{biotac_sim2real}, the image observations have higher variance and more noise. This paper introduces elaborate model designs, data augmentations, and self-supervised adaptations to resolve such difficulties.

Fig.~\ref{fig:architecture} shows the training structure of the network.
The image variational autoencoder (VAE) (in green) reconstructs depth maps $I$ to $\hat{I}$ and is trained as a $\beta$-VAE:
\begin{equation}
\label{img_vae_loss}
\begin{aligned} 
\ell_{I} = \text{MSE} (I-\hat{I}) + \lambda_{I} \text{KL} (q(z_{I}|I) \parallel \mathcal{N}(0,1))
\end{aligned}
\end{equation}
where $q$ is the image encoder, $\lambda_{I}$ is the weight for the KL divergence term, and $z_I$ is the latent vector.

We adopt the convolutional mesh autoencoders (COMA)\cite{coma} for the volumetric mesh VAE (shown in blue). COMA uses spectral graph convolutional networks~\cite{cheb_layer} to extract features and a hierarchical pooling operation to reduce vertices.
The network is trained with:
\begin{equation}
\label{mesh_vae_loss}
\begin{aligned} 
\ell_{M} = \text{MSE} (\mathcal{M}-\hat{\mathcal{M}}) + \lambda_{M} \text{KL} (h(z_{M}|I) \parallel \mathcal{N}(0,1))
\end{aligned}
\end{equation}
where $h$ is the mesh encoder, $\lambda_{M}$ is the KL loss weight, $z_{M}$ is the latent vector, and the MSE is computed based on corresponding vertex positions $(\mathcal{V}, \hat{\mathcal{V}})$.

The latent projection model (shown in orange) is comprised of three fully connected layers. It is trained in a supervised manner with the encoder and decoder frozen.
The details for the network are presented in Section~\ref{subsec: net_arch}. The latent dimensions and weights are chosen via hyperparameter search, which is discussed in Section~\ref{subsubsec: latent_dim_loss}.

\subsection{Self-Supervised Adaptation}
\label{subsec: adapt}
When deploying the trained network to the real world, covariate shift problems may reduce the performance significantly~\cite{cgpn}. Moreover, the real-world data only has depth maps $\{I_{j}\}$, the ground-truth volumetric meshes are not available, making it hard to fine-tune the network in a supervised manner. Thus, we propose a self-supervised adaptation framework (Fig.~\ref{fig:architecture}) to resolve the covariate shift.

Specifically, the reconstructed volumetric mesh $\hat{\mathcal{M}}$ is rendered to the image $\tilde{I}$ using a differentiable renderer, which allows gradients to propagate backward.
In parallel, we use the pre-trained image VAE to reconstruct the input depth map $\hat{I}$.
The image VAE works as a noise filter as suggested in~\cite{vae_noise}. In practice, removing the image VAE can lead to poor adaptation results, which is demonstrated in Section~\ref{subsubsec: ablation}. The network is adapted using the mesh decoder with frozen weights, to minimize the loss:
\begin{equation}
\label{adapt}
\ell_{adapt} = \text{MSE} (\tilde{I}-\hat{I})
\end{equation}

\subsection{Datasets}
Labeled synthetic data $\{(I_i, \mathcal{M}_i)\}$ and unlabeled real-world data $\{(I_j)\}$ are required to train the image-to-mesh projection and adapt the network among different domains.

\subsubsection{Synthetic Data}
\label{subsubsec: syn_data}

Labeled image-mesh pairs $\{(I_i, \mathcal{M}_i)\}$ for $i\in [1, ..., N]$ can be simulated using FEM and synthetic cameras. In this work, FEM is performed using the GPU-based Isaac Gym~\cite{isaac_gym}. Isaac Gym models the dynamics of deformable bodies using linear-elastic models and assumes isotropic Coulomb contacts. 
The results of the simulation are optimized to match the real-world deformation. 

\begin{figure}[tb]
\begin{center}
	\includegraphics[width=3.0in]{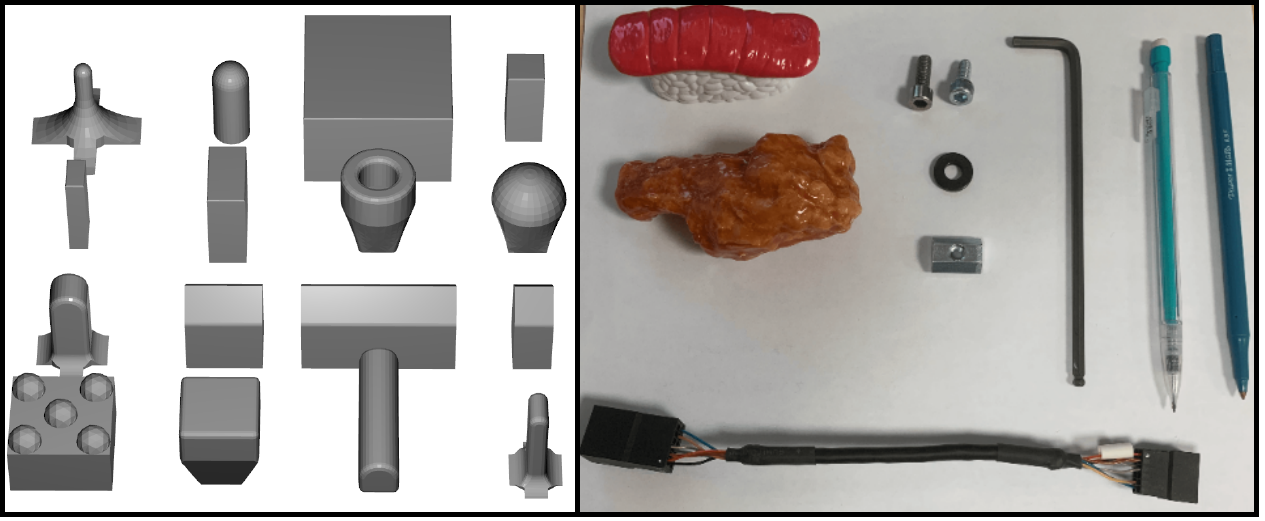}
	\caption{\textit{Left}: Primitive indenters in simulation. \textit{Right}: Novel contact objects in the real world.}
	\label{fig: indenters}
\end{center}
\end{figure}

A FEM model for the GelSlim is created with a similar procedure as~\cite{biotac_sim2real}. The elastomer pad is modeled as a cylinder with a 1.75\textit{cm} radius and 0.3\textit{cm} height. The volumetric mesh has 5,415 nodes and 23,801 edges. 
A rigid backplate is added to imitate the structure of the physical GelSlim, Fig~\ref{fig: teaser}(b).
To generate labeled data pairs, 16 primitive indenters (Fig.~\ref{fig: indenters}--Left) are utilized to interact with the elastomer at randomized positions and rotations. 
The primitive shapes contain a variety of complexity, texture, and geometry to reflect daily household objects.

\begin{figure}[tb]
\begin{center}
	\includegraphics[width=3.4in]{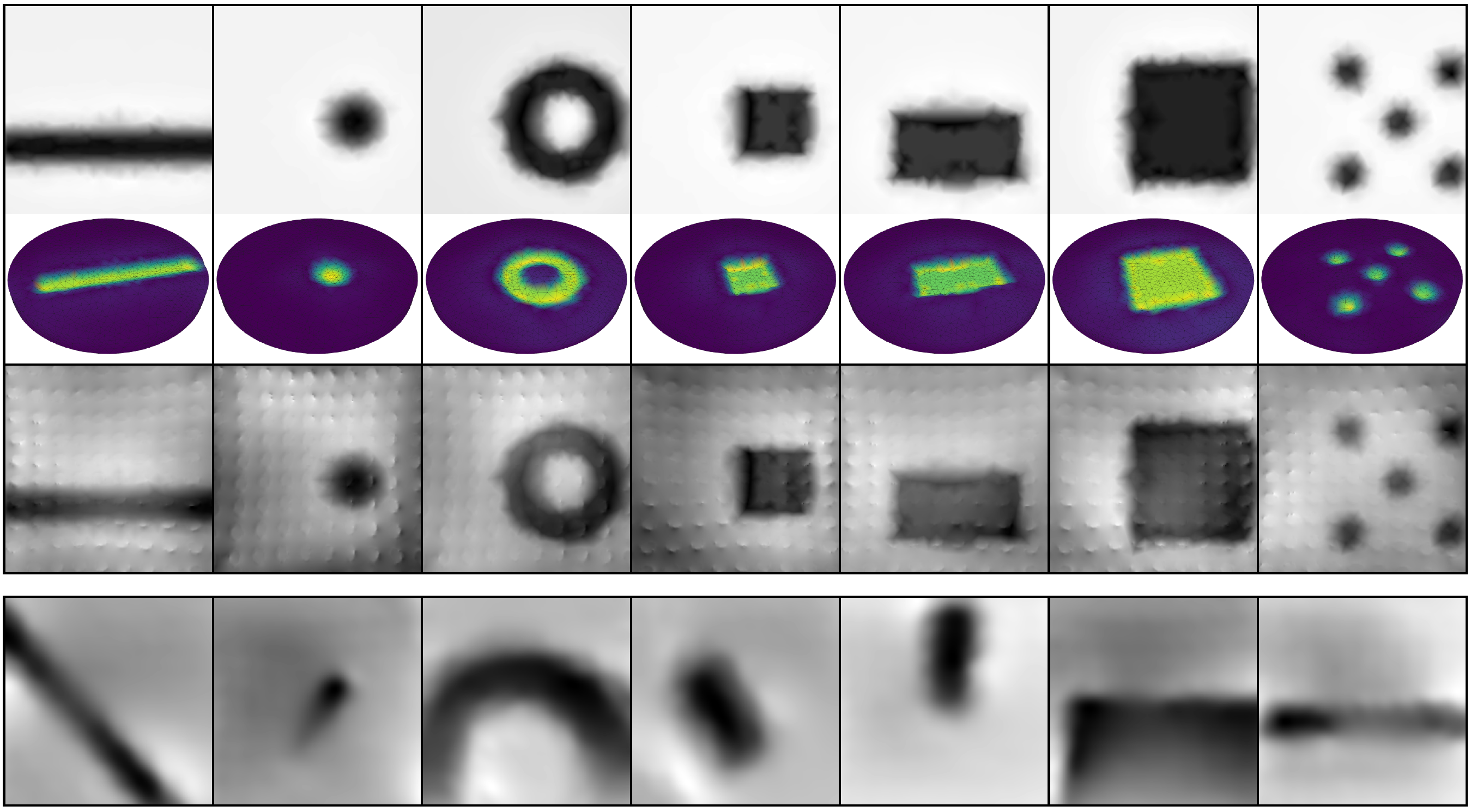}
	\caption{Data samples. \textit{Top}: Raw synthetic depth observations, corresponding ground-truth meshes, and augmented synthetic depth observations. \textit{Bottom}: Real-world depth observations for sample indenters.}
	\label{fig: data_sample}
\end{center}
\end{figure}

The Isaac Gym simulator collects vertex positions $\mathcal{M}$ at each contact trajectory.
The depth map $I$ is then rendered based on the mesh $\mathcal{M}$ with a synthetic camera. This paper uses an orthographic camera with a $\pm$1.75\textit{cm} view range, which aligns with the specifications of the physical GelSlim. To optimize the FEM model in Isaac Gym, this paper reuses the calibration data in~\ref{subsubsec: img_obs}, contact images of a marble of known dimensions.
From the depth map $I$, the contact position can be accurately estimated by finding the maximum displacement point. The contact trajectory can then be reproduced in the simulation, which yields a deformed mesh $\mathcal{M}$. Then, a depth image $\tilde{I}$ is rendered based on the simulated mesh $\mathcal{M}$. 
The elastic modulus $E$, Poisson’s ratio $\nu$, and surface friction $\mu$ are designated as free parameters in the simulator. A cross-entropy search strategy is used to find the best parameters:
\begin{equation*} 
E,\nu,\mu = \arg \min_{E,\nu,\mu} \left \| I - \tilde{I} \right \|
\end{equation*}
The optimal values for $E,\nu,\mu$ are 145MPa, 0.32, 0.94, respectively.
Fig.~\ref{fig: data_sample} shows examples of synthetic data pairs with the calibrated FEM model.

\subsubsection{Real-World Data}
\label{subsubsec: real_data}
Real-world datasets $\{ I_j \}$ are obtained with physical GelSlim sensors and various indenters (Fig.~\ref{fig: data_sample}).
Primitive indenters are 3D printed and interaction with the sensor is randomized.
Besides primitive shapes, several household and industrial objects are used as a novel set (Fig.~\ref{fig: indenters}--Right).
The novel set represents common objects that the GelSlim will work with.
Moreover, we use two GelSlim sensors to collect real-world data.

\subsubsection{Image Augmentations}
\label{subsubsec: aug_data}
As shown in Fig.~\ref{fig: data_sample}, the appearance of synthetic images is quite different from that of real-world depth maps.
The depth reconstruction process for the physical GelSlim introduces significant noise into the image, enlarging the sim-to-real gap.
To enhance the performance in the real world, this paper injects Perlin noise and adds a real-world reference noise image into the synthetic images~\cite{cgpn}.
The Perlin noise provides a realistic gradient for the image and imitates the real-world camera noise. The reference image provides sensor-specific noise.
Fig.~\ref{fig: data_sample} provides examples of the noised images.

In total, 1.28M unique labeled image-mesh pairs were obtained from the simulator, and 1,651 real-world images were obtained for 2 GelSlim sensors with 19 indenters.


\section{Experiments \& Results}
\label{sec: exp_results}
In this section, we present the network details, experiments for supervised image-to-mesh projection, self-supervised adaptation, and a comparative evaluation with a baseline.

\subsection{Network Details}
\label{subsec: net_arch}
As described in Section~\ref{subsec: i2m} and~\ref{subsec: adapt}, we use an image VAE, a mesh VAE, and a latent projection module. In the image VAE, the encoder includes five downsampling layers with feature sizes 32, 64, 128, 256, 512 and two fully connected layers with 128 neurons each. 
In the volumetric mesh VAE, the encoder consists of four Chebyshev convolutional filters~\cite{cheb_layer} with feature sizes 16, 16, 16, 32 and an output fully connected layer with 128 neurons. Each Chebyshev convolution is down-sampled by a factor of four.  
The image and mesh decoder are symmetric with the encoders. The latent projection module has three fully connected layers with 256, 512, and 256 neurons. All networks use the Adam optimizer with a learning rate of $1e-3$ and decay of $0.99$.

\subsection{Supervised Projection}
\label{subsec: super_project}

Our proposed supervised image-to-mesh projection depends on several hyperparameters. In this section we empirically estimate these. Furthermore, we pre-train the VAEs prior to training the image-to-mesh projection. We evaluate the pre-training by comparing with training the image-to-mesh projection directly from scratch.   

The results reported here use a $80/20$ split on the synthetic dataset for training and validation. Each model was trained for 300 epochs. We report the mean validation root-mean-square-error (RMSE) for the projected meshes. 

\subsubsection{Latent Dimensions} 
\label{subsubsec: latent_dim_loss}

\begin{table}
\centering
\caption{Experiments with synthetic data pairs. The root-mean-square error (RMSE, in \textit{cm}) is measured between the ground-truth vertex positions $\mathcal{M}$ and predicted vertex positions $\hat{\mathcal{M}}$.
(a) the results with different dimensions of latent space, (b) the results with different loss weights.
}
\label{tab1}

\subcaptionbox{}{\begin{tabular}[t]{c|ccc}
\textbf{\backslashbox{$d(z_I)$}{$d(z_M)$}}    & 64     & 128             & 256    \\ \hline \hline
64        & 0.221 & 0.152          & 0.200 \\
128       & 0.232 & \textbf{0.141} & 0.192 \\
256       & 0.210 & 0.167          & 0.189  
\end{tabular}}

\subcaptionbox{}{\begin{tabular}[t]{c|cccc}
\textbf{\backslashbox{$\lambda_I$}{$\lambda_M$}}    & 0      & 200             & 400    & 800    \\ \hline \hline
0         & 0.141 & 0.073          & 0.124 & 0.150 \\
100       & 0.082 & \textbf{0.012} & 0.025 & 0.037 \\
200       & 0.094 & 0.035          & 0.031 & 0.046     
\end{tabular}}
\end{table}

We compare the image-to-mesh projection results for a 64, 128, and 256-dimensional latent space for each VAE, shown in Table~\ref{tab1} (a). The 128-dimensional latent space for both VAEs gives the best results.

\subsubsection{Loss Weights}
\begin{figure}[tb]
\begin{center}
	\includegraphics[width=3.4in]{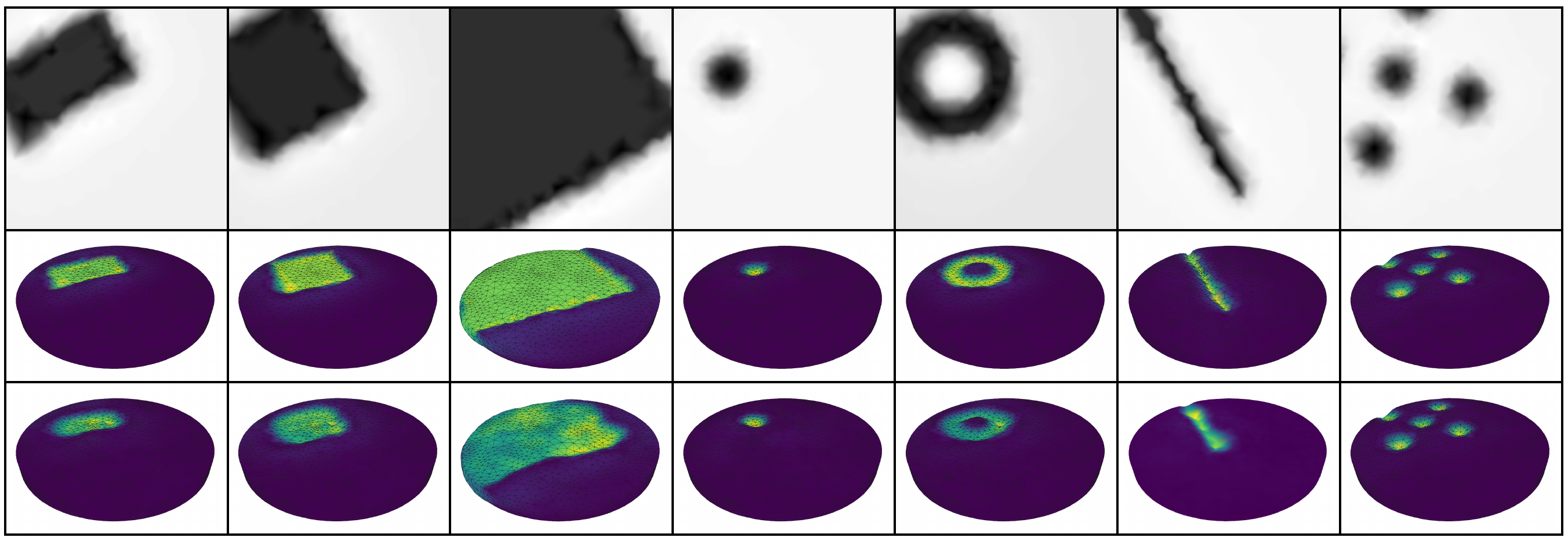}
	\caption{Image-to-mesh projection results with synthetic data. \textit{First row}: Input depth observations. \textit{Second row}: Corresponding ground-truth mesh. \textit{Third row}: Reconstructed volumetric mesh with our approach.}
	\label{fig: img2mesh_results}
\end{center}
\end{figure}

We also compared the effectiveness of different values for $\lambda_I$, $\lambda_M$, from eqs. (\ref{img_vae_loss}) and (\ref{mesh_vae_loss}), shown in Table~\ref{tab1} (b).
We can see that the variational encoding, i.e., $\lambda_I > 0$, $\lambda_M > 0$, significantly improves the performance of latent projection, with best performance for $\lambda_I=100$, $\lambda_M=200$. This suggests that the KL divergence term enforces a more meaningful latent distribution compared to a vanilla autoencoder. Fig.~\ref{fig: img2mesh_results} shows a batch of projection results using the best performing model.


\subsubsection{Pre-training}
Given the best performing model, we investigate the usefulness of the VAE pre-training.
We trained the image-to-mesh network from scratch with variational encoding.
The training and validation errors were 0.009\textit{cm} and 0.085\textit{cm}, respectively. This suggests that the network overfits without the pre-training, which aligns with the findings presented in~\cite{biotac_sim2real}.

\subsection{Self-Supervised Adaptation}
We propose a self-supervised adaptation method and synthetic data augmentations to resolve the covariate shift problem, as discussed in~\ref{subsec: adapt} and~\ref{subsubsec: aug_data}.
This section provides results and ablation studies for the proposed method. We show that neither adaptation nor augmentation can achieve the objective alone, and the image VAE improves the adaptation results. Finally, we demonstrate that the proposed methods can adapt networks from simulation to reality, from primitive to novel contacts, and from one sensor to another.


\begin{figure}[tb]
\begin{center}
	\includegraphics[width=2.91in]{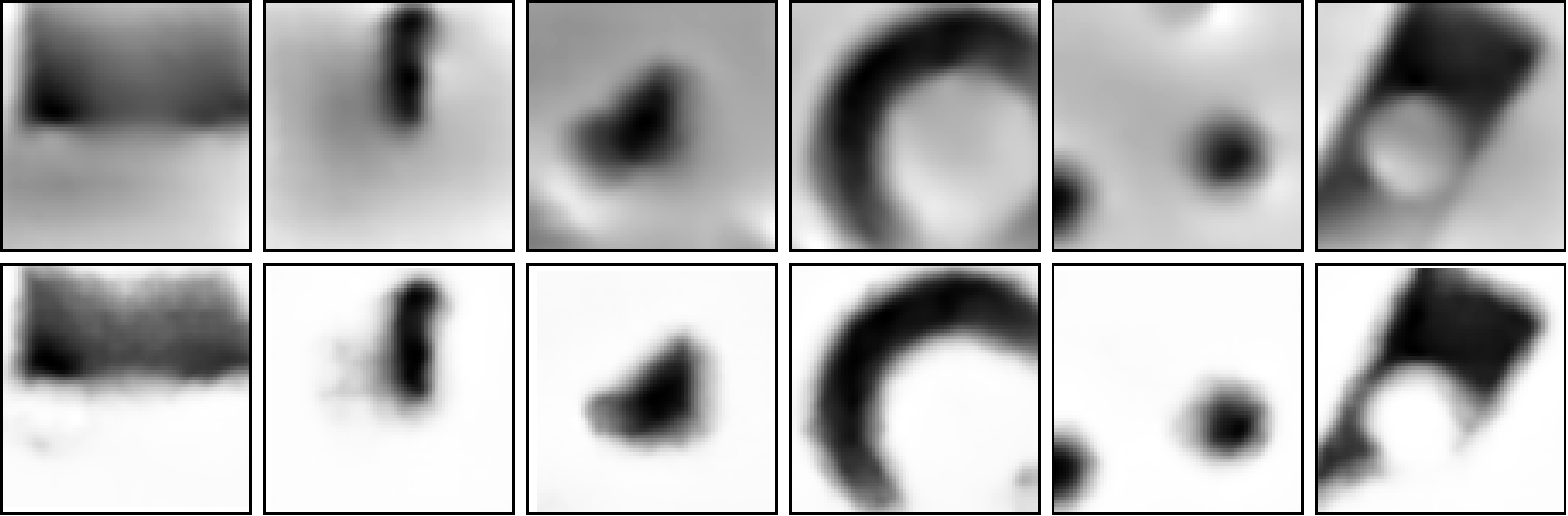}
	\caption{Reconstruction results for the image VAE with real-world images. \textit{First row}: Real-world image observations. \textit{Second row}: Reconstructed image with pre-trained VAE. The image VAE can effectively remove visual noises for both primitive and novel contacts.}
	\label{fig: vae_results}
\end{center}
\end{figure}

The adaptation is performed with the real-world dataset $\{ I_j \}$, without ground-truth mesh availability. To evaluate the performance of the adaptations, we use the RMSE between $\hat{I}$ and $\tilde{I}$ as the evaluation metric, where $\hat{I}$ is the reconstructed input depth map via the pre-trained image VAE. As shown in Fig.~\ref{fig: vae_results}, we can observe that the image VAE is robust in different domains and can effectively remove noise. Specifically, we tested the VAE on augmented synthetic images. Results show that the pre-trained image VAE can reconstruct the clean depth map with a RMSE of 0.07\textit{cm}. 

\subsubsection{Ablation Studies}
\label{subsubsec: ablation}

\begin{table}
\centering
\caption{Experiments with real-world data. The root-mean-square error (RMSE) is measured between reconstructed images $\tilde{I}$ and rendered images $\hat{I}$. 
(a) ablation studies for adaptation, data augmentation, and VAE filtering. (b) domain adaptation results.}
\label{tab2}

\subcaptionbox{}{
\begin{tabular}[t]{l|c}
                     & RMSE (\textit{cm})          \\ \hline \hline
Adapt + Aug.         & \textbf{0.12}     \\
No Aug. No Adapt     & 1.03              \\ \hline
Only Aug.            & 0.57              \\
Only Adapt           & 0.79              \\ \hline
Adapt + Aug. w/o VAE & 0.87              
\end{tabular}}

\subcaptionbox{}{
\begin{tabular}{c|c}
Source $\rightarrow$ Target & RMSE before/after Adaptation (\textit{cm}) \\ \hline \hline
\textsl{Sim-Prim.} $\rightarrow$ \textsl{Real-Prim}   & 0.57 $\rightarrow$ 0.12     \\
\textsl{Sim-Prim} $\rightarrow$ \textsl{Real-Prim-2} & 0.77 $\rightarrow$ 0.20     \\
\textsl{Real-Prim} $\rightarrow$ \textsl{Real-Prim-2} & 0.35 $\rightarrow$ 0.16     \\ \hline
\textsl{Real-Prim} $\rightarrow$ \textsl{Real-Novel}   & 0.64 $\rightarrow$ 0.41     \\ \hline
\textsl{Sim-Prim} $\rightarrow$ \textsl{Real-Novel}   & 1.30 $\rightarrow$ 0.62     
\end{tabular}}
Networks were trained or tuned on source domains and then adapted to target domains. The RMSEs were measured before and after the adaptation.
\end{table}

\begin{figure}[tb]
\begin{center}
	\includegraphics[width=3.4in]{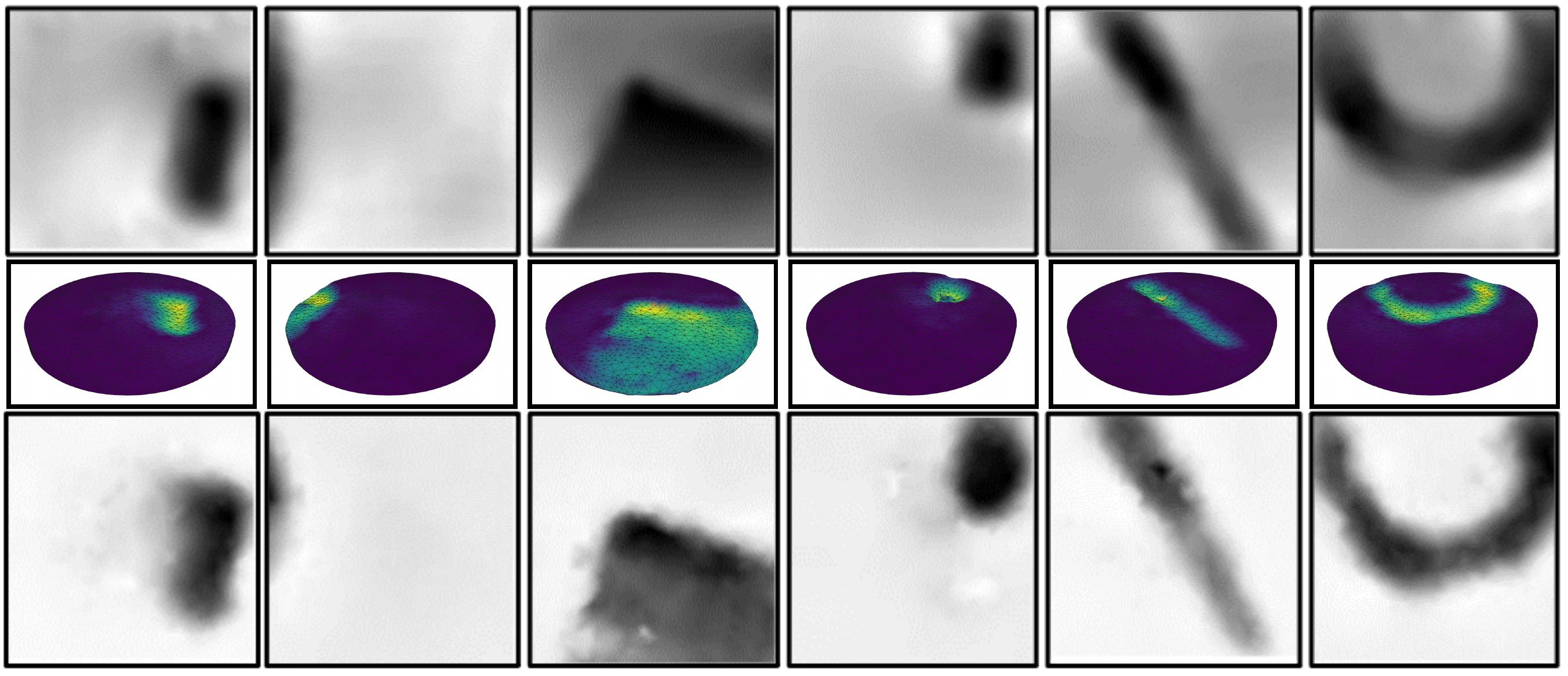}
	\caption{Experiments with real-world primitive contact objects. \textit{First row}: Input depth observations. \textit{Second row}: Reconstructed volumetric meshes. \textit{Third row}: Rendered depth images from reconstructed meshes. }
	\label{fig: real_prim}
\end{center}
\end{figure}

We compare the effects of the adaption model, synthetic data augmentations, and image VAE filtering. The results are listed in Table~\ref{tab2} (a). As the table shows, 
the data augmentation and self-supervised adaptation both contribute to resolving the sim-to-real gap. We observe that using only adaptation, or only augmentation, results in lower performance. The reason for higher performance when both are combined is two-fold. 
On one hand, the data augmentation enlarges the distribution of the synthetic dataset, which causes the real-world data to be within distribution (or close to).
On the other hand, the adaptation model transfers the network from the simulated distribution to the real-world distribution, ensuring invariant feature encodings.
Table~\ref{tab2} (a) also shows that the VAE filter improves adaptation performance. It removes visual noises in real-world data and stabilizes the adaptation process. A batch of qualitative reconstruction examples is shown in Fig~\ref{fig: real_prim}.

\subsubsection{Domain Adaptations}

\begin{figure}[tb]
\begin{center}
	\includegraphics[width=3.4in]{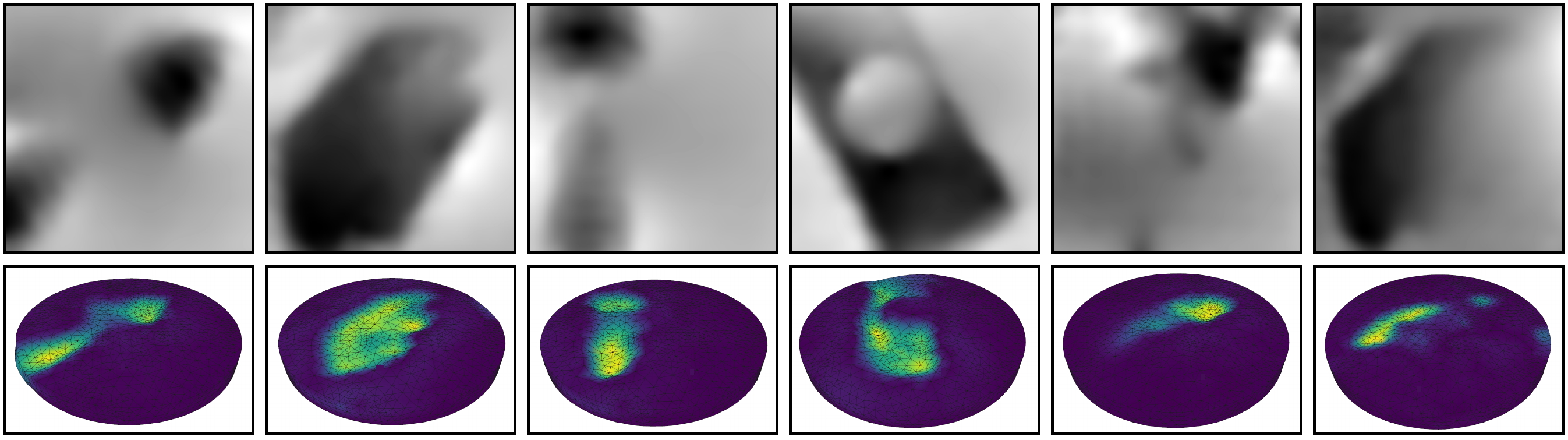}
	\caption{Experiments with real-world novel contact objects. 
	\textit{First row}: Input depth observations. \textit{Second row}: Reconstructed volumetric mesh from the network.}
	\label{fig: novel_gs1_results}
\end{center}
\end{figure}

Sections~\ref{subsubsec: syn_data} and~\ref{subsubsec: real_data} introduce various data domains, including simulated data with primitive contact objects (\textsl{Sim-Prim}), real-world data with primitive contact objects (\textsl{Real-Prim}), real-world data with novel contact objects (\textsl{Real-Novel}), and real-world primitive data with a second GelSlim sensor (\textsl{Real-Prim-2}).

While we showed the performance of the \textsl{Sim-Prim} $\rightarrow$ \textsl{Real-Prim} experiment above, Table~\ref{tab2} (b) and Fig.~\ref{fig: novel_gs1_results} show the transfer results among other domains.
The networks were first pre-trained or fine-tuned on source domains and then adapted to target domains. 
Experiments \textsl{Sim-Prim} $\rightarrow$ \textsl{Real-Prim}, \textsl{Sim-Prim} $\rightarrow$ \textsl{Real-Prim-2}, and \textsl{Real-Prim} $\rightarrow$ \textsl{Real-Prim-2} were executed with the same primitive shapes. The adaptation improves performance in all cases. For experiment \textsl{Real-Prim} $\rightarrow$ \textsl{Real-Novel}, acquisition was done with the real sensor, but adaptation now is for primitive to novel shapes. From Table~\ref{tab2} (b) we see that while the performance improves, improvement is less compared to the prior experiments. For the final experiment \textsl{Sim-Prim} $\rightarrow$ \textsl{Real-Novel} transfer is both from sim-to-real, as well as from primitive to novel shapes, and thus is hardest. Again, adaptation significantly improves performance and predicted deformations were visually accurate (see Fig.~\ref{fig: novel_gs1_results}). The results suggest that the proposed adaptation method can effectively improve the performance of the network under both visual noise and shape differences.

Overall performance for experiment \textsl{Sim-Prim} $\rightarrow$ \textsl{Real-Novel} is less compared to the other experiments. The covariate shifts for visual noise and shape differences are not correlated, and adaptation for each separately performs better compared to adaptation for both. Further optimizing performance for both in a self-supervised manner is a challenging topic for future work. 



\subsection{Baseline Comparisons}
\label{subsec: baseline_comp}
For regression from image observations to mesh deformations, two methods were evaluated: 1) our proposed method, denoted as \textit{Volumetric Mesh}, 
and 2) a surface reconstruction baseline~\cite{gs2,gs3}, denoted as \textit{Surface Mesh}. The latter uses tracking markers to determine the movement of the elastomer surface. Note that the \textit{Surface Mesh} method does not estimate the volumetric mesh directly, but rather gives a sparse surface deformation field for each contact. 

\begin{figure}[tb]
\begin{center}
	\includegraphics[width=3.0in]{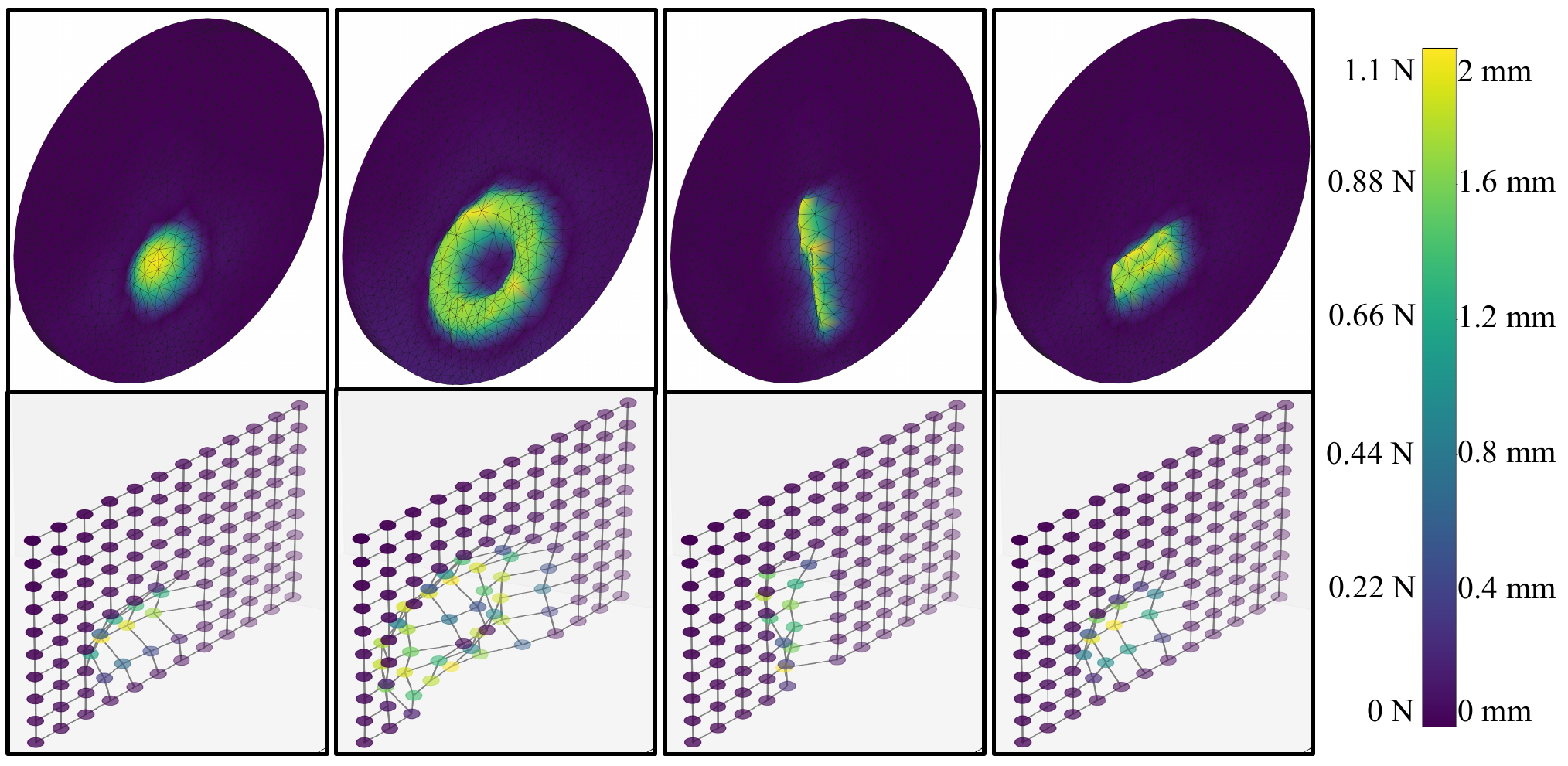}
	\caption{
	\textit{First row}: Reconstructed meshes and estimated contact forces with the proposed approach, \textit{Volumetric Mesh}. \textit{Second row}: Comparison with baseline method, \textit{Surface Mesh}~\cite{gs3}.}
	\label{fig: force_reconstruct}
\end{center}
\end{figure}

Fig.~\ref{fig: force_reconstruct} shows the reconstructed meshes with both methods. Interestingly, the computation takes $0.02$ \textit{sec}. for the \textit{Volumetric Mesh} synthesis with our proposed approach versus $0.04$ \textit{sec}. for the \textit{Surface Mesh} method (potentially due to the requirement of marker detection). Fig.~\ref{fig: force_reconstruct} shows correspondences between the \textit{Volumetric Mesh} and the \textit{Surface Mesh} on the elastomer surface. 
In addition, we also conducted contact force estimations of the GelSlim based on the predicted meshes. An inverse FEM was used to compute the contact force with a linear-elastic model~\cite{gs3}.
Compared to the \textit{Surface Mesh} method, our method constructed more plausible and denser force distributions with the volumetric FEM mesh (Fig.~\ref{fig: force_reconstruct}). For example, predictions around contact edges were more realistic and had higher resolution. Predicted force profiles were also smoother, which was due to the influence of internal vertices. We hypothesize that such denser force distributions obtained from our method may help improve policy learning for robotic manipulation tasks. 




\section{Discussion and Conclusion}
\label{sec: discussion}
This paper presents a framework to synthesize volumetric meshes of vision-based tactile sensor for novel contact interactions.
Our work has several key contributions. 
First, we present a 3D FEM simulator for vision-based tactile sensors and a simulator calibration approach. Second, we generate a dataset for the GelSlim sensor with both simulated and real-world contacts using primitive and novel shapes. Third, we propose a label-free adaptation method and image augmentations for domain transfers; we show that this approach can effectively transfer networks to various visual and different shape scenarios. Lastly, our network efficiently reconstructs the volumetric mesh with depth images and precisely estimates the contact profiles of different shapes. Using these learned and adapted networks, our method can reconstruct the deformations of the elastomer for vision-based tactile sensors in various domains, as indicated by the quantitative and qualitative results.

The present work also has some limitations. First and foremost, although volumetric meshes can obtain dense force estimation and contact patch reconstruction, we do not explicitly demonstrate the application of tactile volumetric meshes on a robotic task. Instead, the focus of this work is on the synthesis of volumetric meshes from tactile imprints. Second, the current network cannot predict the dynamics of the elastomer, which may be prohibitive for performing model predictive control applications, as these require valid prediction of futures. 
In our future work, we will focus on addressing these limitations and develop volumetric mesh-based techniques for extensive robotic manipulation tasks, such as rope manipulation, peg-in-hole insertion, extrinsic contact estimation, and contact prediction.



\addtolength{\textheight}{-1cm}   



\bibliographystyle{IEEEtran}
\typeout{}
\bibliography{main}

\end{document}